\documentclass[conference]{IEEEtran}
\IEEEoverridecommandlockouts
\usepackage{cite}
\usepackage{amsmath,amssymb,amsfonts}
\usepackage{algorithm}
\usepackage{graphicx}
\usepackage{textcomp}
\usepackage{xcolor}
\usepackage{multirow}
\usepackage{multicol}
\usepackage{algpseudocode}

\usepackage{caption}
\usepackage{subcaption}

\usepackage{tikz}
\usepackage{textcomp}
\usepackage[hidelinks]{hyperref}

\newcommand\copyrighttext{%
  \footnotesize \textcopyright 2021 IEEE. Personal use of this material is permitted.
  Permission from IEEE must be obtained for all other uses, in any current or future
  media, including reprinting/republishing this material for advertising or promotional
  purposes, creating new collective works, for resale or redistribution to servers or
  lists, or reuse of any copyrighted component of this work in other works.
  DOI: \href{https://doi.org/10.1109/BigComp51126.2021.00035}{10.1109/BigComp51126.2021.00035}}
\newcommand\copyrightnotice{%
\begin{tikzpicture}[remember picture,overlay]
\node[anchor=south,yshift=10] at (current page.south) {\fbox{\parbox{\dimexpr\textwidth-\fboxsep-\fboxrule\relax}{\copyrighttext}}};
\end{tikzpicture}%
}

\def\BibTeX{{\rm B\kern-.05em{\sc i\kern-.025em b}\kern-.08em
    T\kern-.1667em\lower.7ex\hbox{E}\kern-.125emX}}

\begin{document}

\algnewcommand{\algorithmicgoto}{\textbf{go to}}%
\algnewcommand{\Goto}[0]{\algorithmicgoto~}%

\title{Attention on Personalized Clinical Decision Support System: Federated Learning Approach
\thanks{This work was partially supported by Institute of Information \& communications Technology Planning \& Evaluation (IITP) grant funded by the Korea government(MSIT) (No.2019-0-01287, Evolvable Deep Learning Model Generation Platform for Edge Computing) and by Institute of Information \& Communications Technology Planning \& Evaluation(IITP) grant funded by the Korea government(MSIT)(2020-0-00364, Development of Neural Processing Unit and Application systems for enhancing AI based automobile communication technology). *Dr CS Hong is the corresponding author.}
}

\author{\IEEEauthorblockN{Chu Myaet Thwal, Kyi Thar, Ye Lin Tun, Choong Seon Hong}
	\IEEEauthorblockA{\textit{Department of Computer Science and Engineering} \\
		\textit{Kyung Hee University}\\
		Yongin-si, 17104, Republic of Korea\\
		\{chumyaet, kyithar, yelintun, cshong\}@khu.ac.kr}
}

\maketitle
\copyrightnotice
\begin{abstract}
Health management has become a primary problem as new kinds of diseases and complex symptoms are introduced to a rapidly growing modern society.
Building a better and smarter healthcare infrastructure is one of the ultimate goals of a smart city.
To the best of our knowledge, neural network models are already employed to assist healthcare professionals in achieving this goal.
Typically, training a neural network requires a rich amount of data but heterogeneous and vulnerable properties of clinical data introduce a challenge for the traditional centralized network.
Moreover, adding new inputs to a medical database requires re-training an existing model from scratch.
To tackle these challenges, we proposed a deep learning-based clinical decision support system trained and managed under a federated learning paradigm.
We focused on a novel strategy to guarantee the safety of patient privacy and overcome the risk of cyberattacks while enabling large-scale clinical data mining. 
As a result, we can leverage rich clinical data for training each local neural network without the need for exchanging the confidential data of patients.
Moreover, we implemented the proposed scheme as a sequence-to-sequence model architecture integrating the attention mechanism.
Thus, our objective is to provide a personalized clinical decision support system with evolvable characteristics that can deliver accurate solutions and assist healthcare professionals in medical diagnosing.
\end{abstract}

\begin{IEEEkeywords}
clinical decision support system, healthcare,  artificial intelligence, sequence-to-sequence network, attention mechanism, federated learning
\end{IEEEkeywords}

\section{Introduction}

Remarkable advances in digital and Internet-of-Things (IoT) technologies bring great potential to transform a wide range of cities into smart cities.
With the rise of the smart cities concept, automation strategy, based on a tremendous deployment of IoT devices and software solutions, helps to improve cities’ services and makes more efficient use of their existing infrastructures and assets~\cite{Zanella_2014}.
Smart cities take the advantages of information and communications technology (ICT) to enhance a better quality of life for their citizens by delivering smarter infrastructures in education, transportation, healthcare, economy, and environmental management~\cite{thwal2019edge,thwal2019uav,tun2019edge}.
Thus, the goal of every smart city is to keep a balance in terms of sustainability in economic, environmental, and social impact~\cite{Mohanty_2016}.
Despite the countless benefits of smart city innovations, many challenges need to be focused on, as rapid urbanization introduces new problems to an already complex environment of a highly-populated area.
Among them, health management has become one of the primary issues as new kinds of infectious diseases (i.e., Ebola, plague, COVID-19, etc.) are introducing to a rapidly growing modern society~\cite{who_dons}.
To prevent such disease outbreaks leading to the global pandemic, the authorities keep investigating the medical science and provide better innovations to the health management sector~\cite{who_pandemics}. 


In recent years, artificial intelligence (AI) has contributed numerous kinds of innovations to the smart city domain and gains great attention in many different and popular applications~\cite{Liu_2018}.
AI also plays as a key component in developing better healthcare infrastructures ranging from patient monitoring to clinical data mining.
As many AI algorithms are capable of learning from data, various types of supervised machine learning models can provide AI-based solutions for medical diagnoses and deliver accurate results to healthcare professionals~\cite{Carchiolo_2019}.
A sufficient amount of clinical data is an important prerequisite for accelerating and promoting healthcare services through AI technology.
For training a better medical diagnosis system, clinical data such as electronic health records (EHRs) from different sources of the medical organization need to be collectively organized~\cite{Dash_2019}.
However, these data may involve not only the privacy of individual patients but also the experimental details of medical organizations.
Due to the heterogeneous and vulnerable properties, dealing with limited clinical data becomes a huge obstacle for big data analysis to facilitate precision medicine while developing better healthcare infrastructures~\cite{xu2019federated}.

\begin{figure*}[t]
	\centering
	\includegraphics[width=\linewidth]{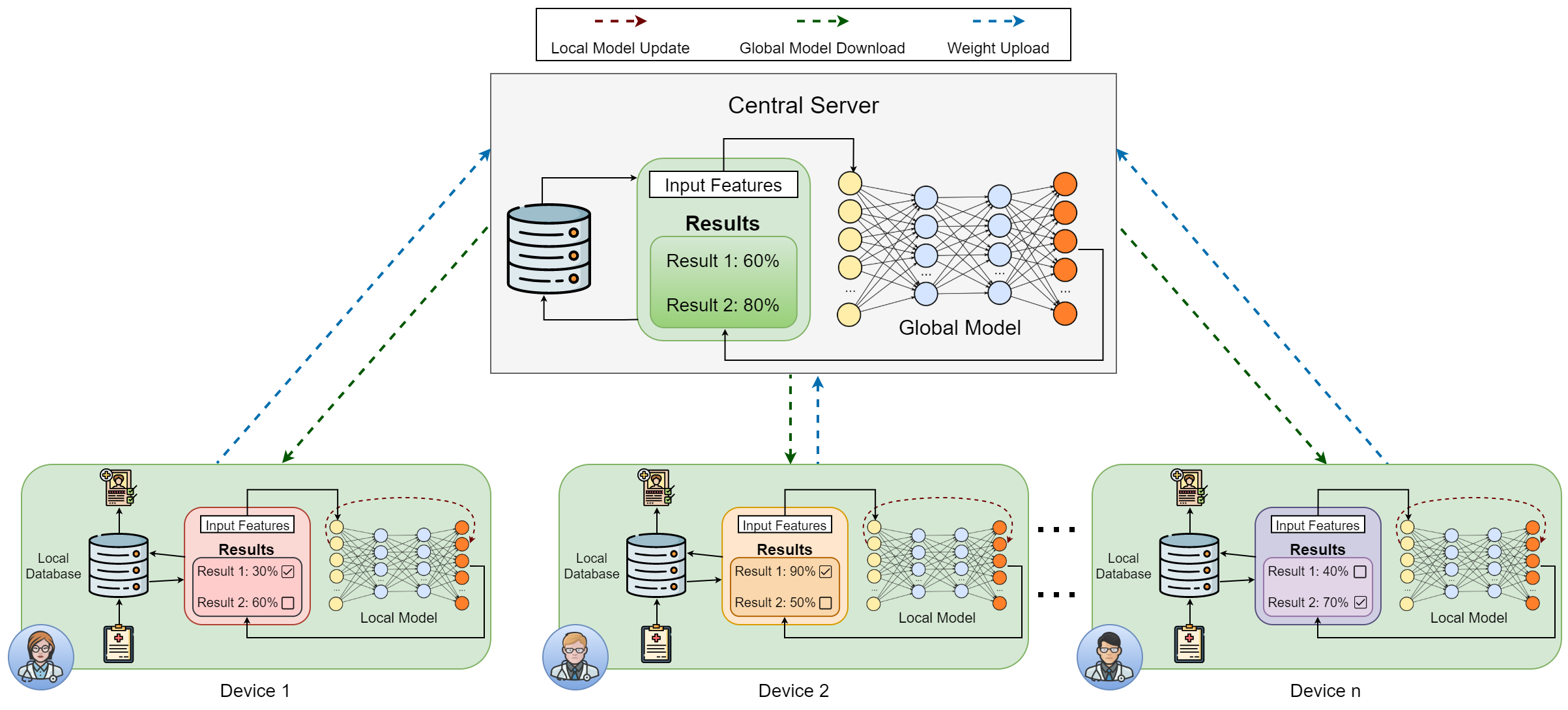}
	\caption{System overview.}
	\label{fig1:system_overview}
\end{figure*}

In this paper, we proposed a deep learning-based clinical decision support system that can provide healthcare professionals with accurate predictions of potential diseases according to patients’ symptoms in their medical records.
We integrate the aid of edge AI that allows the processing of AI algorithms locally on edge devices (i.e., healthcare professionals’ personal computers or mobile phones) for reducing the cost of data communication to perform the real-time operations and avoiding the transmission of vulnerable clinical data for a privacy perspective~\cite{Lee_2018}.
We exploit the federated learning paradigm that can work with a certain degree of non-IID (Independent and Identically Distributed) data from heterogeneous sources by learning a shared global model at the central server while keeping all the sensitive clinical data on each local device~\cite{Li_2020}.
As a result, local models can benefit from a global model with a predictive capacity of training on a rich source of clinical data while not compromising their data as well as preventing the cause of misdiagnosis.

Moreover, most of the existing medical diagnosis models need to be re-trained from scratch whenever new input features (i.e., symptoms) and target labels (i.e., diseases) are added in order to expand the prediction for new diseases~\cite{Brisimi_2018, choudhury2019}.
In our proposed system, we designed a sequence-to-sequence network for training local medical diagnosis models in which symptoms and diseases are treated as input and output sequences, respectively.
For the models to be able to focus more on the relevant input sequences and predict accurate output sequences, we adopt the attention mechanism in designing the network architecture~\cite{Luong_2015}.
Thus, our system can easily adapt new input features and target labels and evolve to predict new kinds of diseases from heterogeneous sources without re-training the models from scratch~\cite{tharevolvable}.

Our objective is to provide an evolvable clinical decision support system that can assist healthcare professionals in the process of medical diagnosis while preserving the patients’ privacy. Hence, patients can also be provided with higher quality, safer, and more efficient healthcare services. The main contributions of this paper can be listed as follows:

\begin{itemize}
	\item We design the system architecture of an edge-based personalized clinical decision support system to jointly work with AI technology.
	\item We apply the attention mechanism in sequence-to-sequence network of our medical diagnosis model for each edge device.
	\item We exploit the federated learning framework to train our personalized models collaboratively in a distributed manner on  the datasets of non-IID symptom-disease network.
	\item We adopt a centralized global model to capture the shared knowledge of all personalized models by aggregating their gradients, regardless of their private information.
	\item We also design our models with evolvable characteristics to be able to adapt to new kinds of symptoms and diseases without re-training from scratch.
\end{itemize}

The rest of this paper is organized as follows: Section II explains the system architecture and methodologies of the proposed clinical decision support system. Section III presents the experiment settings for training the models. Simulation results are discussed in Section IV. Finally, Section V presents our conclusions and future works.

\section{System Architecture}

Let us consider a personalized healthcare architecture in which different medical diagnosis models jointly participate in learning the heterogeneous clinical data from the EHRs of regional healthcare professionals.
The global model in the centralized server does not take part in the training process but captures the shared knowledge of all local models and monitors their training rounds.
Each local model trains its own neural networks on its private training dataset of medical records. Each trained model is deployed onto the  personal devices of individual healthcare professional to assist in classifying the diseases based on the related symptoms.
Classification outputs from those models are analyzed by the healthcare professionals based on their expert knowledge to evaluate the final result.
The evaluated results of symptom-disease pairs are stored in the local database for the model calibration.
The local model sends its model updates to the central server after each round of the training process.
The server aggregates the collected local model updates and creates a new single global model. Then, the aggregated global model is distributed for replacing the local models and updating their performance.
Fig.~\ref{fig1:system_overview} shows an overview of our proposed clinical decision support system operating under the federated learning paradigm.

\begin{figure*}[t]
	\centering
	\includegraphics[width=\linewidth]{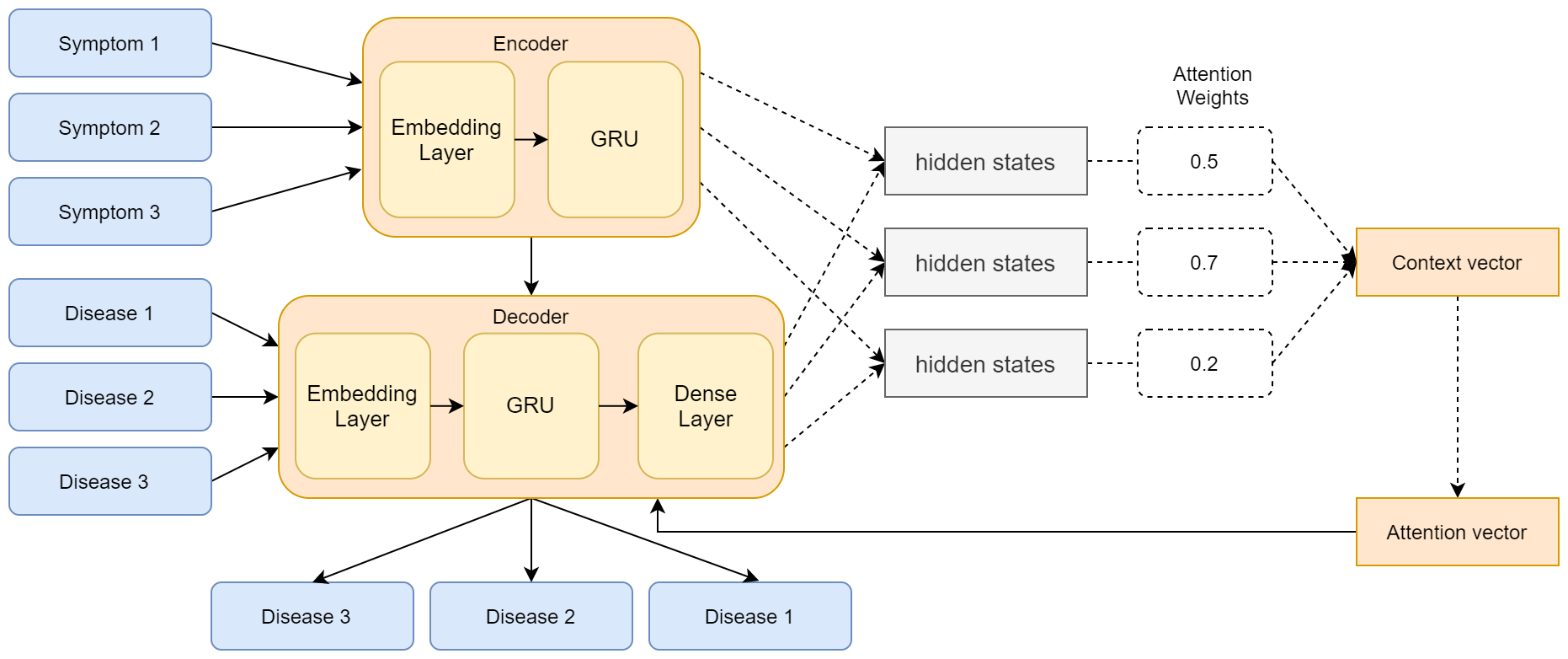}
	\caption{Attention network architecture.}
	\label{fig2:attention_network}
\end{figure*}

\subsection{Attention on Edge AI-based Medical Diagnosis Model}

We design the architecture of our proposed clinical decision support system as a deep recurrent neural network (Deep RNN).
While constructing the local medical diagnosis models, we process the sequence-to-sequence modeling for symptom-disease mapping, where symptoms are input sequences as well as diseases are treated as output sequences.
Each mapping is comprised of two RNN models for the encoder network and decoder network, respectively.
We establish direct shortcuts between the input symptoms and the target diseases by applying the attention mechanism in the encoder-decoder network.
The encoder network with hidden states $h_s$ processes the input sequence $x$ and each of the source hidden state $\bar{h}_s$ is compared with the current target hidden state $h_t$ by a score function in Bahdanau's additive style~\cite{bahdanau2014neural} which can be defined as
\begin{equation}
score(h_t, \bar{h}_s) = v_a^{\top} \tanh(W_1h_t+W_2\bar{h}_s),
\label{eqn:score_function}
\end{equation}
where $W_1$, $W_2$ and $v_a$ are the weight matrices. Then, the result is normalized and the attention weights $\alpha_{ts}$ are produced by
\begin{equation}
\begin{aligned}
\alpha_{ts} = \frac{\exp(score(h_t, \bar{h}_s))}{\sum_{s'=1}^{S}\exp(score(h_t, \bar{h}_{s'}))}.
\label{eqn:attention_weights}
\end{aligned}
\end{equation}
Based on the calculated attention weights, a smaller dimensional representation of the input sequence, which is known as a context vector $c_t$ is computed as the weighted average of the source states and can be defined by 
\begin{equation}
c_t = \sum_{s} \alpha_{ts}\bar{h}_s.
\label{eqn:context vector}
\end{equation}
The decoder network is initialized with the context vector and trained to generate an output sequence $y$.
The context vector is concatenated with the current target hidden state and yields the attention vector $a_t$ which can be derived by
\begin{equation}
a_t = f(c_t, h_t) = \tanh(W_c[c_t;h_t]).
\label{eqn:attention_vector}
\end{equation}
The final attention vector is used to derive the softmax logit and loss.
By processing our symptom-disease mapping problems as sequence-to-sequence models and applying the attention mechanism, our models do not need to be re-trained from scratch whenever new input features are added to the training data.
The architecture of an attention network embedded in each local medical diagnosis model of our proposed system is shown in Fig.~\ref{fig2:attention_network}.

\subsection{Collaboration of Federated Learning}

For a neural network to work well, it is required to be trained on a rich data source.
In the traditional training pipeline of a neural network, the training process is centralized and it becomes infeasible to communicate as the data to be collected grows larger.
To alleviate the expensive communication costs together with security and privacy issues of the centralized training scheme, we collaborate the federated learning settings to our model architecture~\cite{xie2020multicenter}.
In the federated learning paradigm of our proposed system, the centralized server initially distributes a base global model $\mathcal{M}_{global}$ to be employed in each personal device of the healthcare authority as a local model $\mathcal{M}_k$ that is parameterized by the weight $W_k$.
Each device stores the personalized symptom-disease knowledge dataset $\mathcal{D}_k = \{X_k, Y_k\}$ in its local database, where $X_k$ denotes the input features (i.e., symptoms) and $Y_k$ denotes the related target labels (i.e., diseases), respectively.
Each dataset $\mathcal{D}_k$ is used to train the local medical diagnosis model $\mathcal{M}_k: X_k \rightarrow Y_k$. The training procedure for each local model can be defined as
\begin{equation}
W'_k = \underset{W_k} {\arg \min} {L(\mathcal{M}_k, \mathcal{D}_k, W_k)},
\label{eqn:local_model_weight}
\end{equation}
where $W'_k$ denotes the updated weight after training and $L(.)$ is the general loss function of each local model argumented with model structure $\mathcal{M}_k$, training data $\mathcal{D}_k$, and the local weight parameter $W_k$.
At the end of each training round, a local model sends its model updates $W'_k$ back to the server.
The central server aggregates all the weight updates collected from the local models and produces a new global model $\mathcal{M}_{global}$ that is parameterized by the global weight $W_g$.
The aggregation mechanism is known as the weighted average of local model parameters which can be defined as
\begin{equation}
{W}_g = \sum_{k=1}^{m} \frac{|\mathcal{D}_k|}{\sum_{k+1}|\mathcal{D}_{k+1}|},
\label{eqn:global_model_weight}
\end{equation}
where $m$ denotes the number of devices participating in the proposed system.
The aggregated global model is distributed back to the local models after each round of global model updates.
By training our RNN models under the federated learning settings, we can reduce the communication costs between the local models and the server.
As well as the local clients are no longer required to reveal their sensitive data to the central server or between each other.
The aforementioned procedure of the federated learning setting is listed in Algorithm~\ref{algorithm:1}.

\begin{algorithm}[t]
	\caption{Federated learning for global model update}\label{algorithm:1}
	\floatname{algorithm}{Procedure}
	
	
	\begin{algorithmic}[1]
		
		\State {\textbf{Inputs:}
			Local model $M_k = \{M_1, M_2, ..., M_K\}$ with parameter $W_k$, Global model $M_{global}$ with parameter $W_g$, Input feature $X_k$, Target label $Y_k$, Local private datasets $D_k = \{D_1, D_2, ..., D_K\}$, where each $D_k = \{X_k,Y_k\}$}
		
		\State {\textbf{Outputs:} Trained global model $M_{global}$ with updated parameter $W_g$}
		
		\State {\textbf{Initialize:} $W_g = random()$}
		
		\For {communication round $t=1,2,...,T$}
		\For {each client $k=1,2,...,K$ in parallel}
		\State {$W_k \leftarrow W_g$}
		\For {local training step $n=1,2,...,N$}
		\State Train $M_k:X_k \rightarrow Y_k$ with training data $D_k$
		\State Update local model parameter: 
		\begin{equation*}
		W'_k=\underset{W_k} {\arg \min} {L(\mathcal{M}_k, \mathcal{D}_k, W_k)}
		\end{equation*}
		\State return updated local parameter $W'_k$ to the central server
		\EndFor
		\EndFor
		\State Update global model parameter:
		\begin{equation*}
		W_g= \frac{1}{K} \sum_{k=1}^{K} W_k
		\end{equation*}
		\EndFor
		\State return $W_g$	
	\end{algorithmic}
\end{algorithm}

\section{Experiment Settings}
In this section, we analyze the effectiveness of our proposed personalized clinical decision support system by demonstrating a proof-of-concept scenario.

\subsection{Datasets}

We use the manually preprocessed dataset~\cite{aniruddha_tapas} retreated from a Disease-Symptom Knowledge Database~\cite{Wang2008} which is generated by an automated method based on the information in textual discharge summaries of patients at New York Presbyterian Hospital in 2014.
In the original database, the first column shows the types of diseases combined with UMLS codes obtained from MedLEE natural language processing system~\cite{Friedman_2004}.
The second column shows the count of disease occurrence. And, the third column shows the associated symptoms obtained by the statistical methods based on frequencies and co-occurrences.
The original dataset is uncleaned and for the training processes to be more accurate, we use a manually preprocessed dataset that contains both training and testing sets of symptoms and diseases.
The preprocessed dataset is cleaned and extensive, where the associations of symptoms are one hot encoded related to each disease.
It contains 4,920 training samples having 41 unique types of diseases and 132 unique types of symptoms.
The description of the preprocessed dataset is described in Table.~\ref{tab1}, where \# denotes the number.


\begin{table}[h]
	\caption{Statistics of dataset.}
	\begin{center}
	\begin{tabular}{|c|c|}
		\hline
		\multicolumn{2}{|c|}{\multirow{2}{*}{Data Description}}                                                                                                                                                                                                                                                                              \\
		\multicolumn{2}{|c|}{}                                                                                                                                                                                                                                                                                                               \\ \hline
		\multirow{3}{*}{\begin{tabular}[c]{@{}c@{}}\# of unique diseases\\ \# of unique symptoms\end{tabular}}                                                                                                                            & \multirow{3}{*}{\begin{tabular}[c]{@{}c@{}}41\\ 132\end{tabular}}                                \\
		&                                                                                                  \\
		&                                                                                                  \\ \hline
		\multirow{7}{*}{\begin{tabular}[c]{@{}c@{}}\# of clients\\ \# of instances for client 1\\ \# of instances for client 2\\ \# of instances for client 3\\ \# of instances for client 4\\ \# of instances for client 5\end{tabular}} & \multirow{7}{*}{\begin{tabular}[c]{@{}c@{}}5\\ 1,000\\ 1,000\\ 1,000\\ 1,000\\ 920\end{tabular}} \\
		&                                                                                                  \\
		&                                                                                                  \\
		&                                                                                                  \\
		&                                                                                                  \\
		&                                                                                                  \\
		&                                                                                                  \\ \hline
	\end{tabular}
\label{tab1}
	\end{center}
\end{table}

We shuffle and split the training data into five subsets regarding local datasets for five clients.
Four of the local datasets contain 1,000 instances and the remaining one contains 920 instances of diseases with related symptoms.
We convert the datasets into text files in which symptoms and diseases are paired as input and output sentences.
For the convenience of using with sequence-to-sequence networks, we add start and end tokens to each sentence and create a word index for mapping symptom to id and a reverse word index for mapping id to disease, respectively.

\begin{figure*}[t]
	\centering
	\begin{subfigure}[t]{0.45\textwidth}
		\centering
		\includegraphics[width=\textwidth]{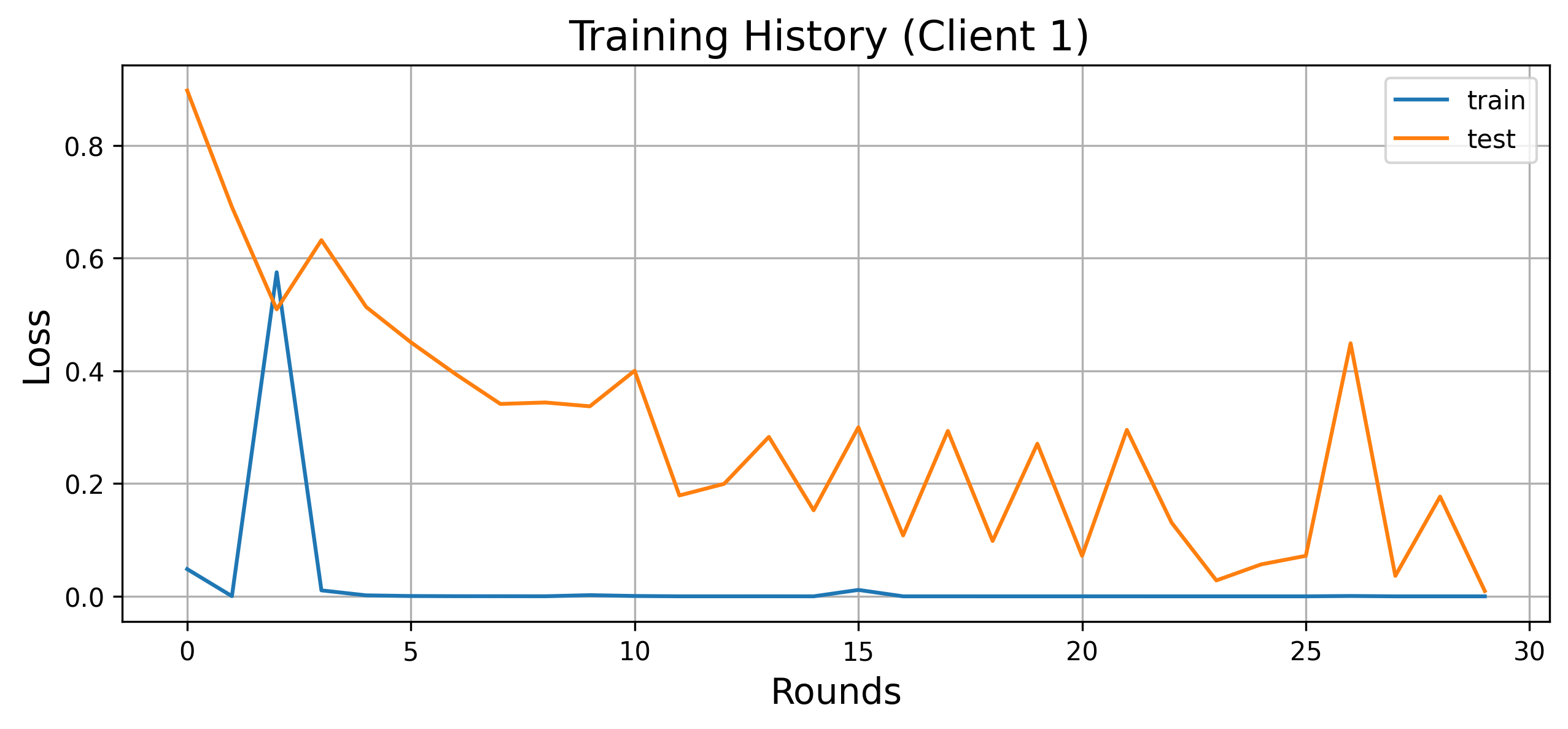}
		\caption{Local model 1}
		\label{fig_a:client_1}
	\end{subfigure}
	\centering
	\begin{subfigure}[t]{0.45\textwidth}
		\centering
		\includegraphics[width=\textwidth]{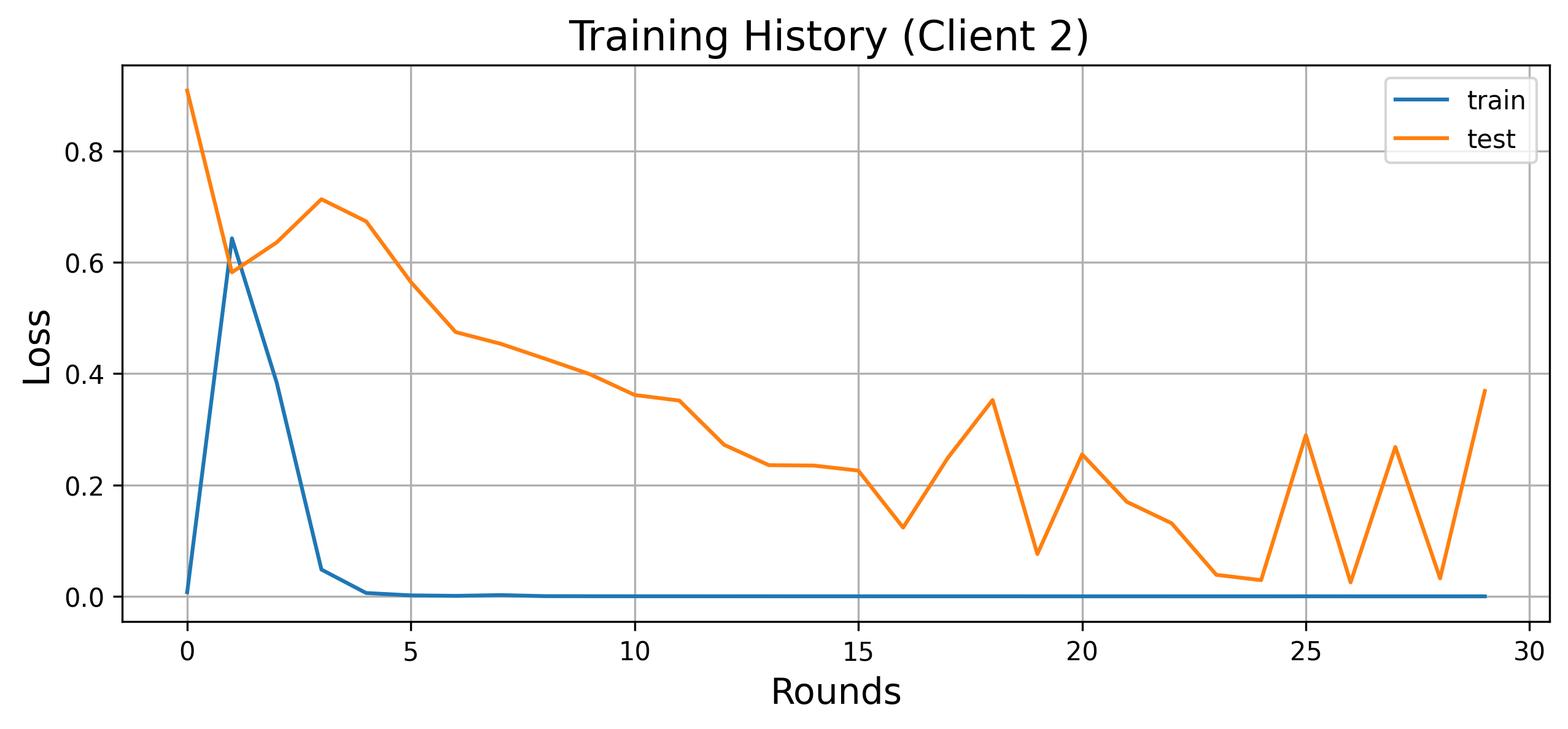}
		\caption{Local model 2}
		\label{fig_b:client_2}
	\end{subfigure}
	\par\medskip
	\centering
	\begin{subfigure}[t]{0.45\textwidth}
		\centering
		\includegraphics[width=\textwidth]{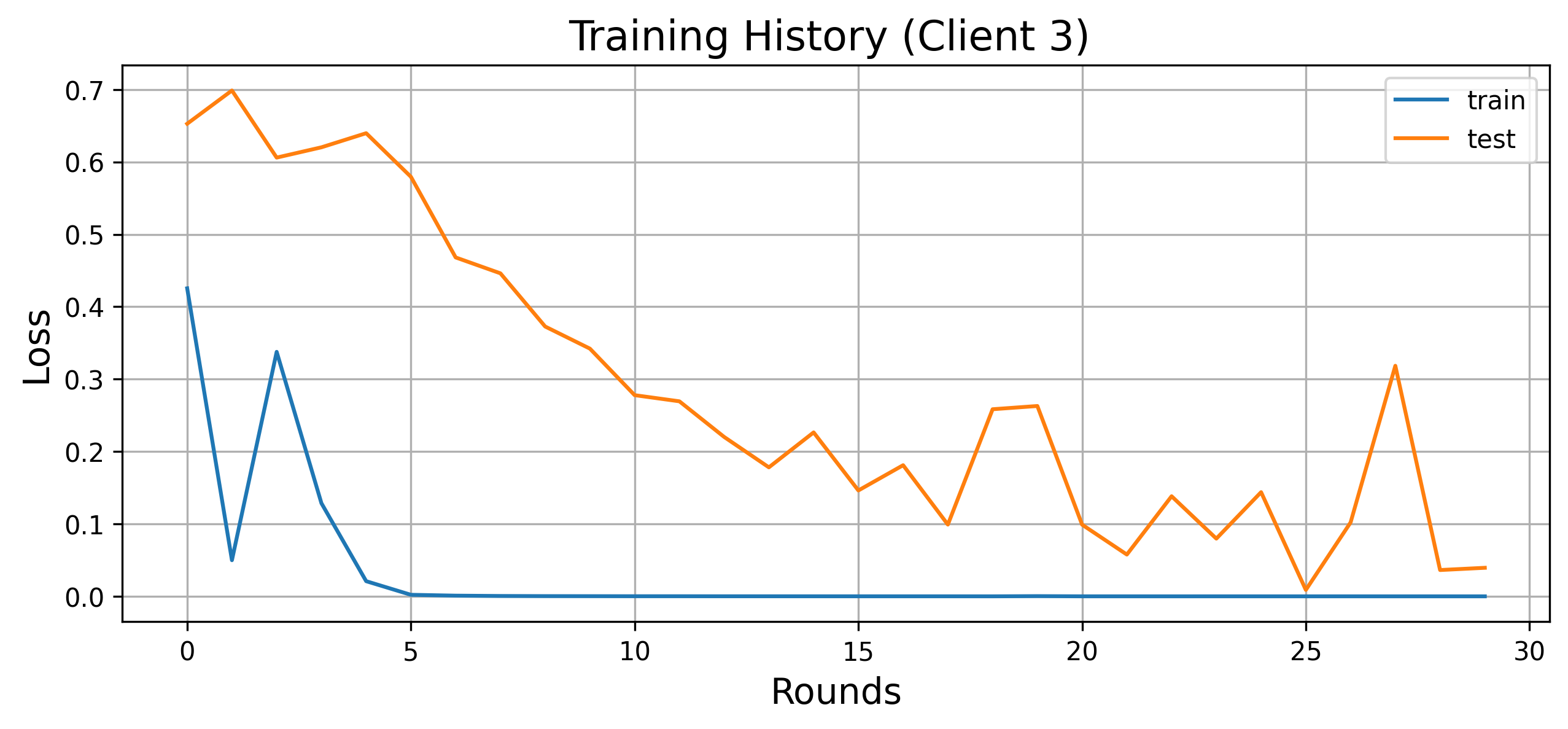}
		\caption{Local model 3}
		\label{fig_c:client_3}
	\end{subfigure}
	\centering
	\begin{subfigure}[t]{0.45\textwidth}
		\centering
		\includegraphics[width=\textwidth]{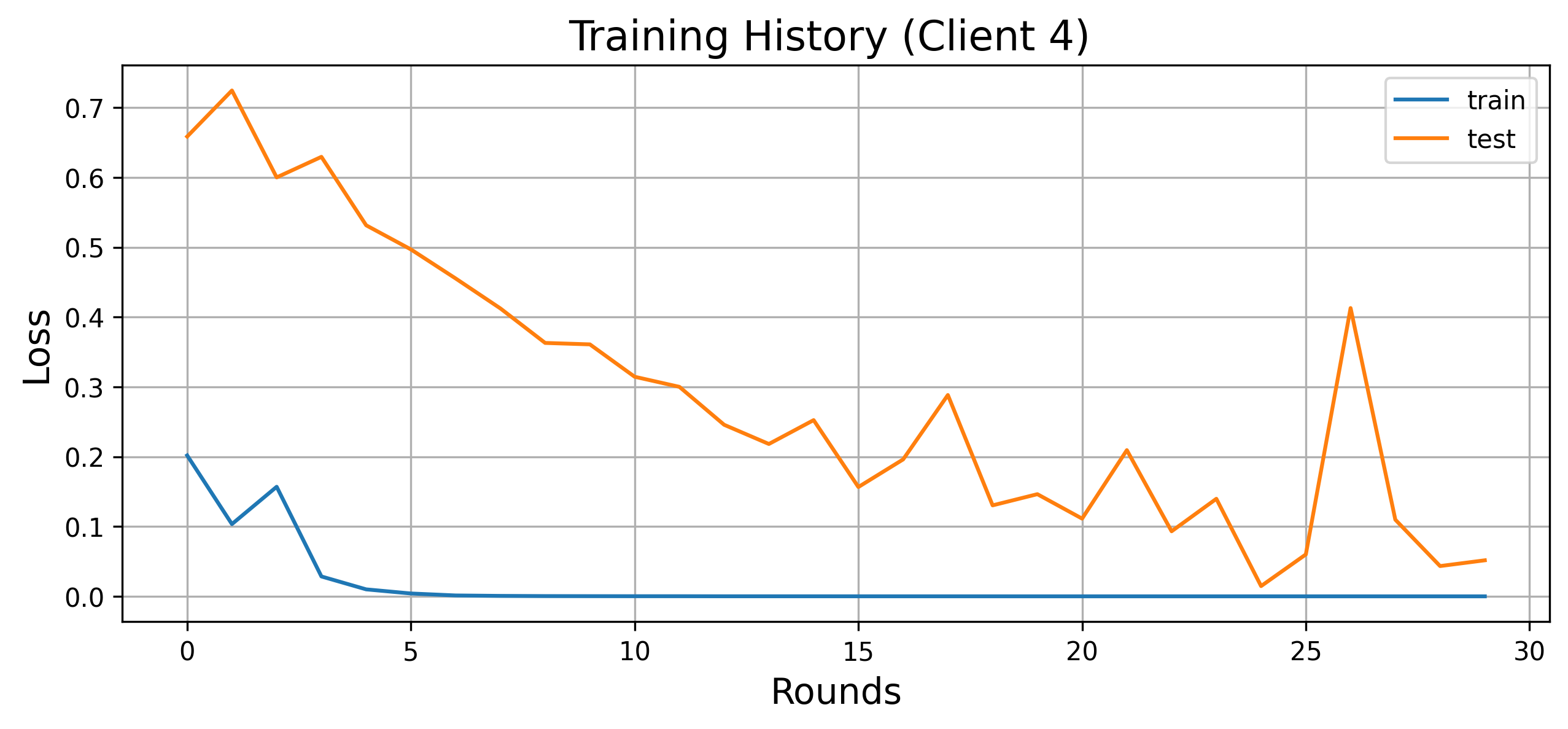}
		\caption{Local model 4}
		\label{fig_d:client_4}
	\end{subfigure}
	\par\medskip
	\centering
	\begin{subfigure}[t]{0.45\textwidth}
		\centering
		\includegraphics[width=\textwidth]{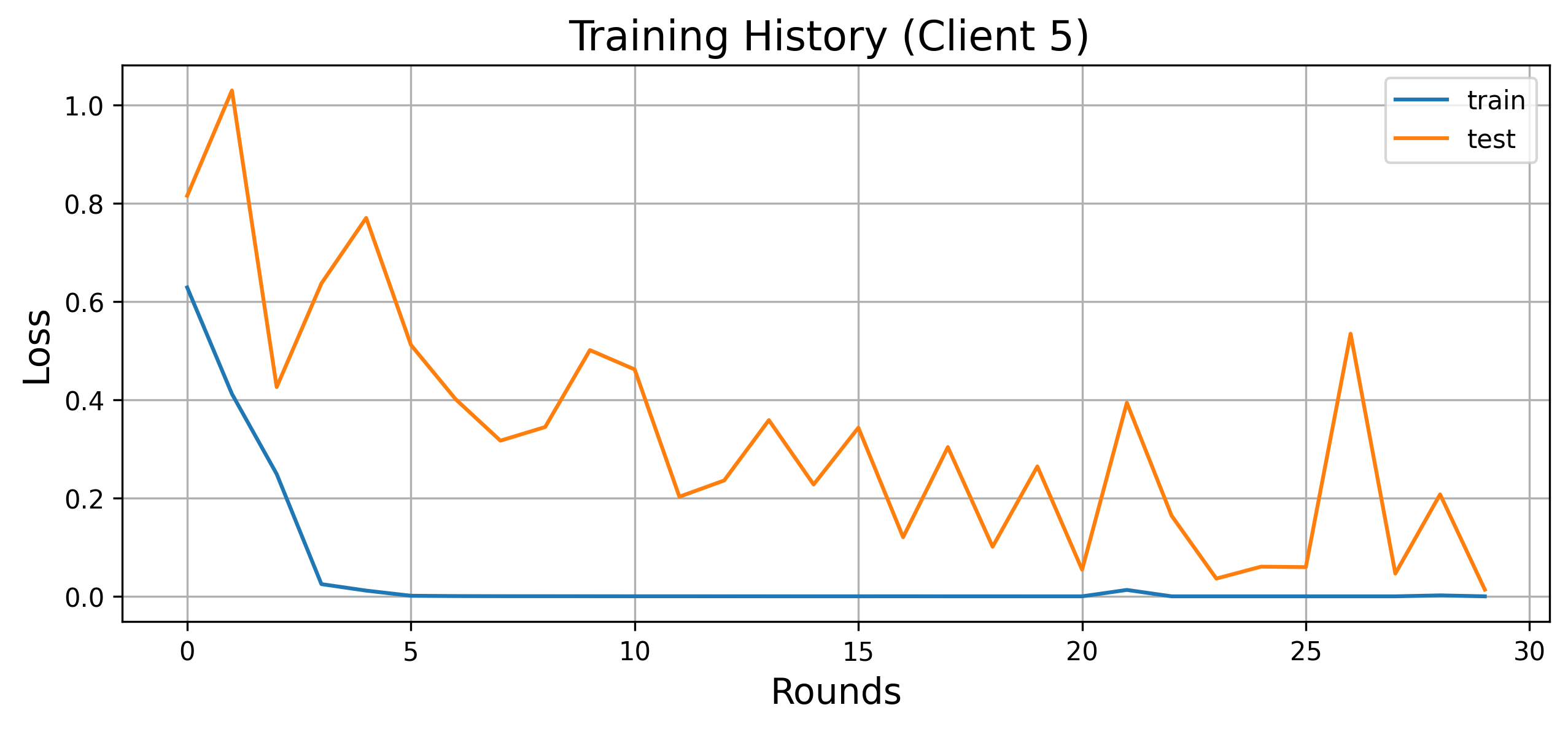}
		\caption{Local model 5}
		\label{fig_e:client_5}
	\end{subfigure}
	\caption{Training history for centralized models.}
	\label{fig3:local_models}
\end{figure*}

\begin{figure*}[t]
	\centering
	\includegraphics[width=150mm]{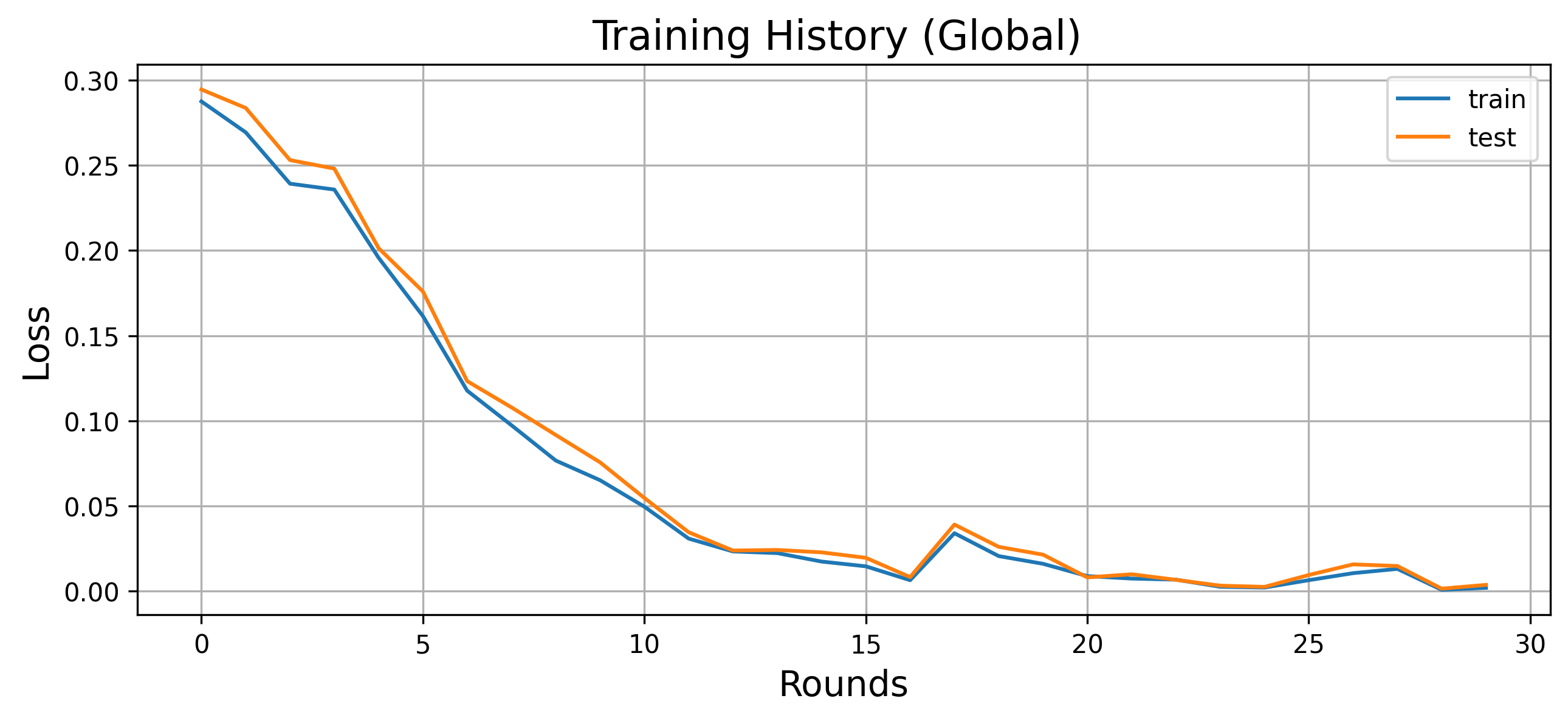}
	\caption{Training history for global model.}
	\label{fig4:global_model}
\end{figure*}

\subsection{Local Models}

We consider the personal devices of healthcare professionals as the clients for local models and construct one RNN model for each client of the local medical diagnosis system.
For constructing the encoder-decoder network, we implement a gated recurrent unit (GRU) with 1,024 cells.
Then, we apply the Bahdanau attention mechanism~\cite{Bahdanau_2016} to our neural networks.

\subsection{Training Settings}

We train each local model on the dataset of the individual client in a distributed manner.
For each client's data, we use 80\% for training and 20\% to validate the model.
We randomize the initialization parameters of models and conduct five times of local training for each client per federated round.
Then, we aggregate the weights of all local models by a weighted averaging mechanism and update the global model's weight.
In the next round of federated training, all the local model weights are updated with the aggregated global model weight.
We perform 30 times of federated training for a better accuracy of the disease prediction system.

\section{Simulation Results}

In this section, we analyze the performance of our proposed clinical decision support system by executing the simulations on the Google Colab GPU backend.
We consider a federated learning-based global healthcare architecture where regional healthcare professionals are participating as local clients with their own datasets.
We develop the medical diagnosis models by using TensorFlow API~\cite{abadi2016tensorflow}.
We consider five clients participating as the local models in our healthcare architecture.
Each local model is trained on the private dataset of an individual client.
We illustrate the training and testing loss of each model as the benchmark centralized approach comparing to our proposed federated learning scheme.


The training history for the centralized models are shown in Fig.~\ref{fig3:local_models}.
As the individual local model is trained on its own local dataset and evaluates the results on the same dataset, all the local models converge after five rounds of the training process.
However, the results of the testing processes are evaluated on the dataset of various clinical data.
Thus, the testing losses of the local models fluctuate as the testing dataset includes unknown data for each local model.
Fig.~\ref{fig_a:client_1} shows the model performance of local client 1 where the training loss converges to the value of 2.1083e-05 and the testing loss varies from 0.0094 to 0.8976 throughout the whole simulation process.
The model performance of local client 2 is shown in Fig.~\ref{fig_b:client_2} where the training loss converges to the value of 2.5834e-05 and the testing loss varies from 0.0250 to 0.9088.
Fig.~\ref{fig_c:client_3} shows the model performance of local client 3 where the training loss converges to the value of 2.0466e-05 and the testing loss varies from 0.0089 to 0.6527.
Fig.~\ref{fig_d:client_4} and Fig.~\ref{fig_e:client_5} show the model performances of local client 4 and local client 5, respectively.
In Fig.~\ref{fig_d:client_4} of local client 4, the training loss converges to the value of 2.1468e-05 and the testing loss varies from 0.0147 to 0.6586.
In Fig.~\ref{fig_e:client_5}, the training loss converges to the value of 2.7095e-05 and the testing loss varies from 0.0092 to 1.2760 throughout the whole simulation process.

Fig.~\ref{fig4:global_model} shows the training history for the global model.
Here, the global model is created with the aggregation of local models and the evaluation results are conducted based on the training data of all clients.
After some communication rounds, as the global model has gained reasonable predictive capacity, its training loss becomes stable and satisfied with the value of 1.4985e-04 at $30^{th}$ communication round.
During the testing process, as the global model is aggregated by all the local models, it has already gained the predictive capacity based on various local datasets and the testing loss is stable and satisfied with the value of 4.1726e-03 at the $30^{th}$ communication round.
According to Fig.~\ref{fig3:local_models} and Fig.~\ref{fig4:global_model}, there is a proof that our proposed federated learning scheme outperforms the centralized approaches in testing with different distributions of clinical data.

\begin{figure}[t]
	\centering
	\includegraphics[width=\linewidth]{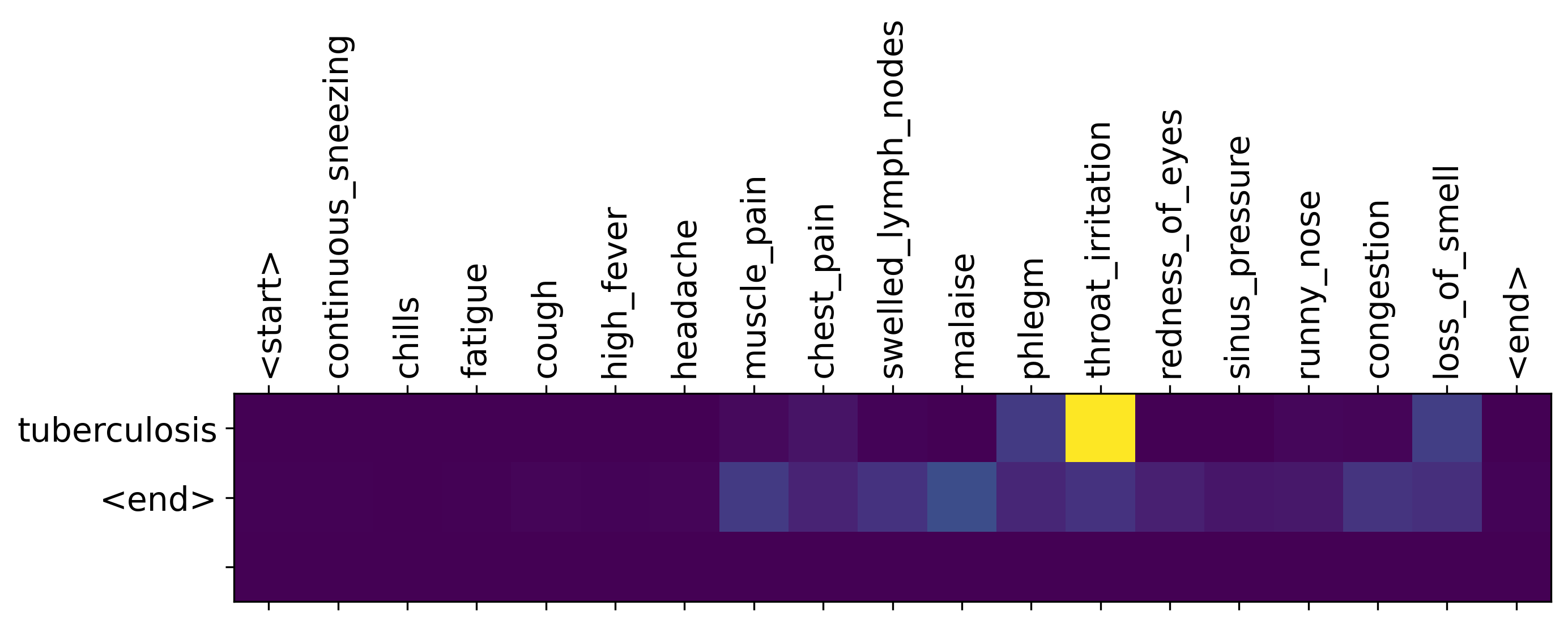}
	\caption{Prediction of tuberculosis.}
	\label{fig5:tuberculosis}
\end{figure}

\begin{figure}[t]
	\centering
	\includegraphics[width=\linewidth]{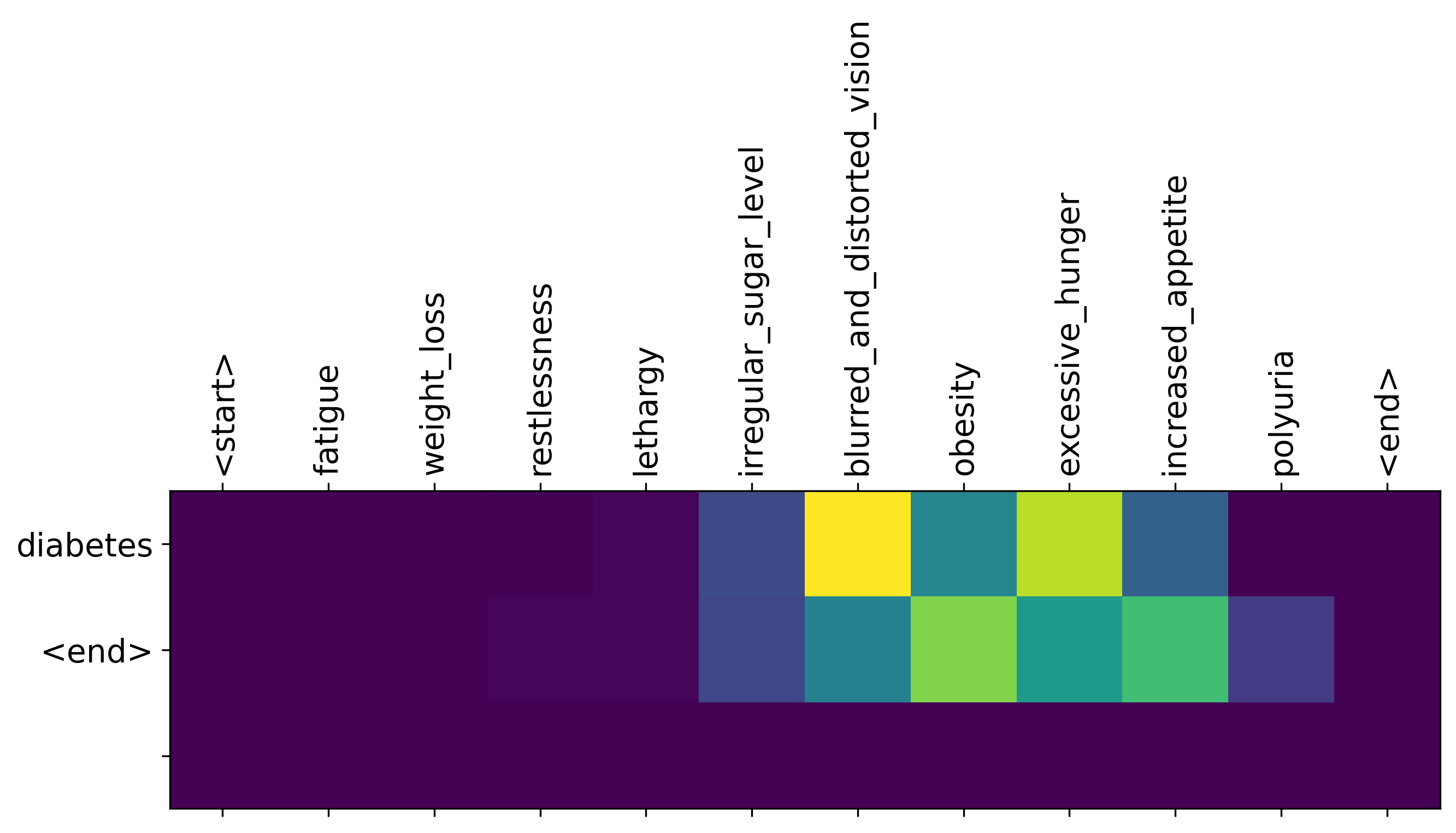}
	\caption{Prediction of diabetes.}
	\label{fig6:diabetes}
\end{figure}

\begin{figure}[t]
	\centering
	\includegraphics[width=\linewidth]{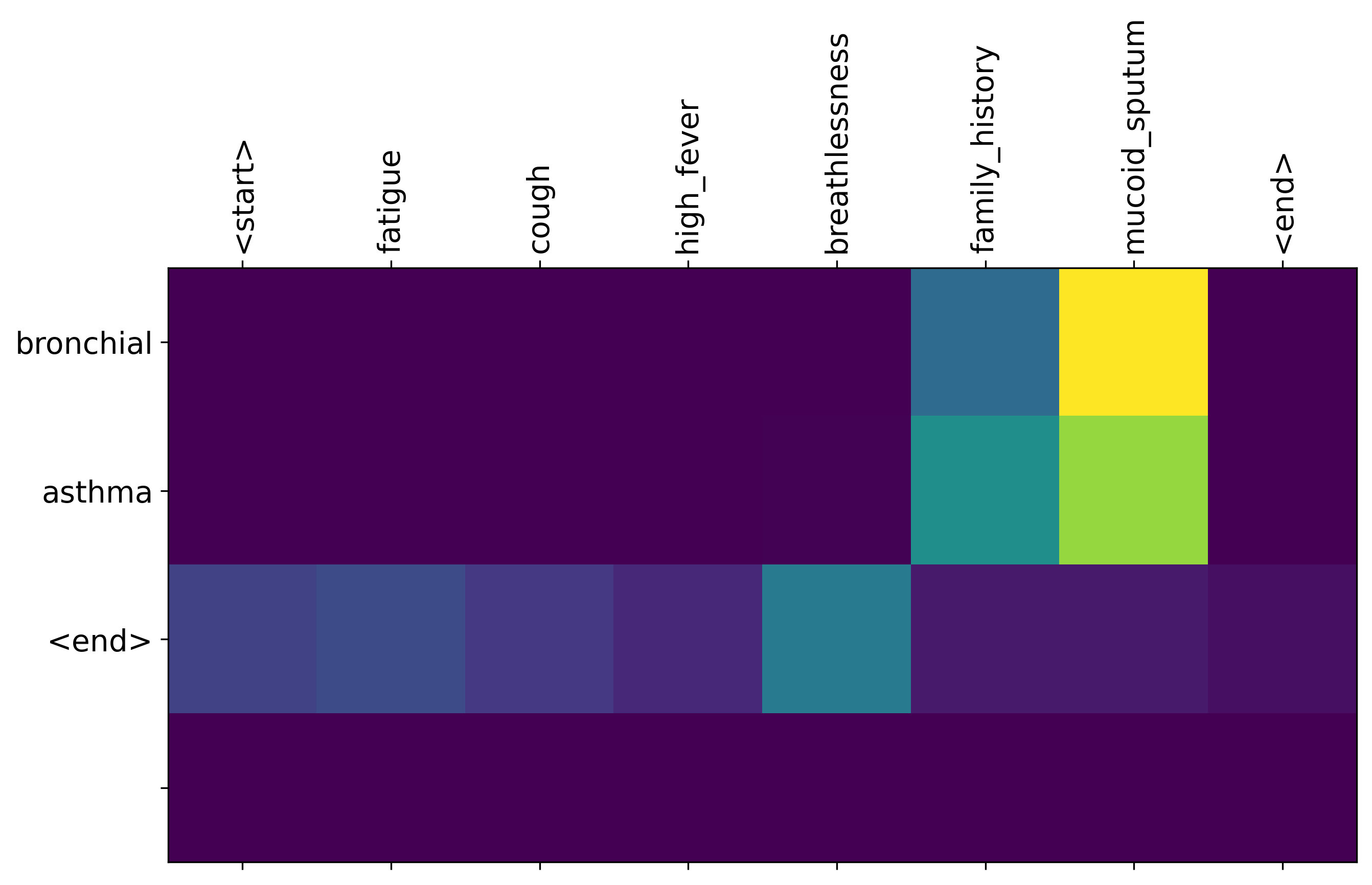}
	\caption{Prediction of bronchial asthma.}
	\label{fig7:bronchial_asthma}
\end{figure}

\begin{figure}[t]
	\centering
	\includegraphics[width=\linewidth]{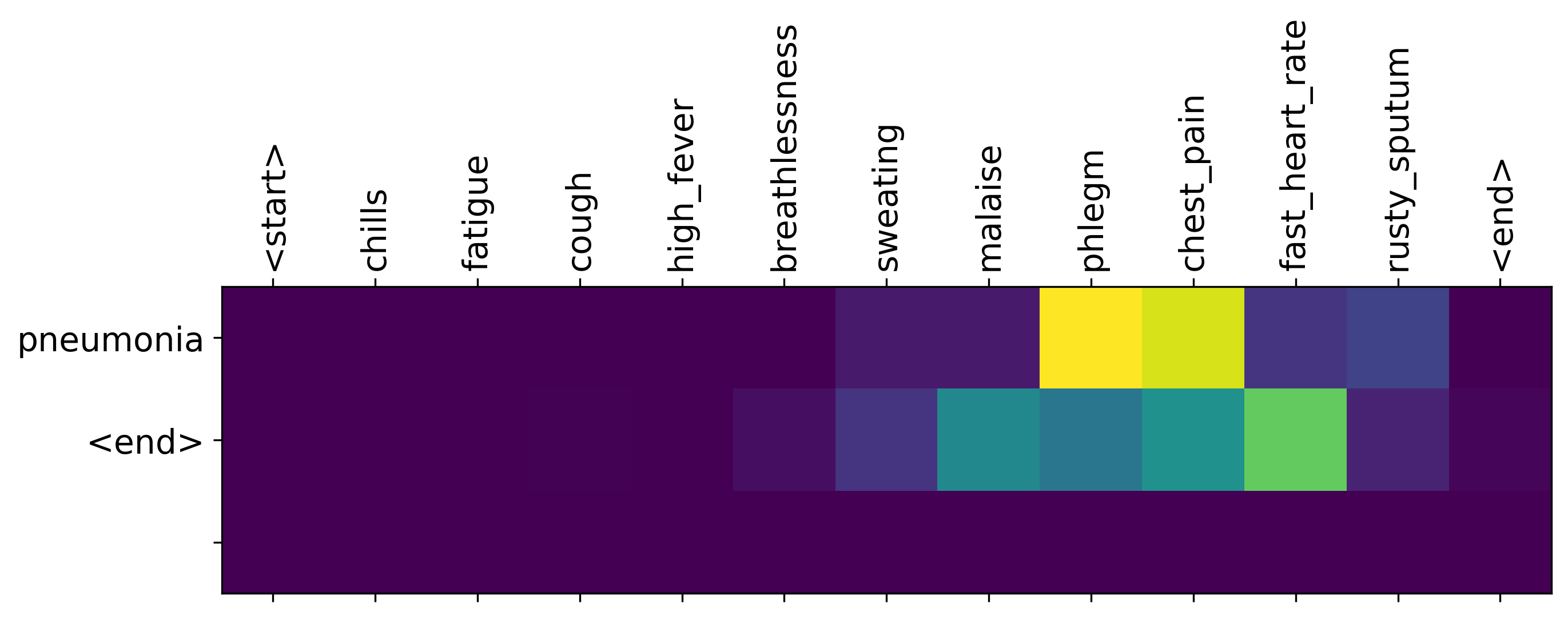}
	\caption{Prediction of pneumonia.}
	\label{fig8:pneumonia}
\end{figure}

The sample prediction results evaluated on the testing dataset are shown in Fig.~\ref{fig5:tuberculosis}, \ref{fig6:diabetes}, \ref{fig7:bronchial_asthma}, and \ref{fig8:pneumonia}, where each prediction represents tuberculosis, diabetes, bronchial asthma, and pneumonia, respectively.
As a result of utilizing the attention mechanism in sequence-to-sequence network architecture, the prediction results of diseases are shown with attention weights, where the most relevant symptom is represented by the yellow color.

\section{Conclusions}

In this paper, we proposed a deep learning-based personalized clinical decision support system trained and managed under a federated learning paradigm to assist healthcare professionals in medical diagnosing.
We integrated the aid of edge AI to train the local models distributively on the personal devices of healthcare professionals for reducing the data communication cost and avoiding the transmission of vulnerable data.
We exploited the federated learning framework to train our personalized models collaboratively by learning a shared global model at the central server while keeping all the privacy of clinical data on each local device.
As a result, our proposed clinical decision support system outperforms the benchmark centralized scheme by enabling large-scale clinical data mining while preserving the sensitive information of patients and medical organizations.
Moreover, we adopted the attention mechanism in designing our medical diagnosis models as sequence-to-sequence networks.
Thus, our system possesses the evolvable characteristics that can easily adapt new input symptoms and expand the prediction for new diseases.
For a more detailed scenario as optimizing the communication costs of the federated model to account for the noisiness of various biomedical data is considered as our future work.

\bibliographystyle{IEEEtran}
\bibliography{references}

\end{document}